\documentclass[conference]{IEEEtran}
\usepackage[utf8]{inputenc}
\IEEEoverridecommandlockouts
\usepackage{cite}
\usepackage{amsmath,amssymb,amsfonts}
\usepackage{algorithmic}
\usepackage{graphicx}
\usepackage{textcomp}

\usepackage{color}
\usepackage{amsthm}
\usepackage{bm,xcolor} % Math packages
\usepackage{mathtools}
\usepackage{bbm}
\usepackage{wrapfig}
\usepackage{subcaption}

\usepackage{booktabs}

\usepackage[colorlinks,citecolor=cyan]{hyperref}
\usepackage{url}
\usepackage{cleveref}

\newtheorem{definition}{Definition}

\newtheorem{remark}{Remark}
\newtheorem{theorem}{Theorem}
\newtheorem{lemma}{Lemma}

\begin{document}

\title{A Convergence Rate for Manifold Neural Networks
\thanks{Accepted to Sampling Theory and Applications Conference 2023, to appear in IEEE Xplore.}
}

\author{\IEEEauthorblockN{Joyce A. Chew}
\IEEEauthorblockA{\textit{Department of Mathematics} \\
\textit{University of California, Los Angeles}\\
Los Angeles, CA 90095, USA \\
joycechew@math.ucla.edu}
\and
\IEEEauthorblockN{Deanna Needell}
\IEEEauthorblockA{\textit{Department of Mathematics} \\
\textit{University of California, Los Angeles}\\
Los Angeles, CA 90095, USA \\
deanna@math.ucla.edu}
\and
\IEEEauthorblockN{Michael Perlmutter}
\IEEEauthorblockA{\textit{Department of Mathematics} \\
\textit{Boise State University}\\
Boise, ID 83725, USA \\
michael.aaron.perlmutter@gmail.com}
}

\maketitle

\begin{abstract}
       High-dimensional data arises in numerous applications, and the rapidly developing field of geometric deep learning seeks to develop neural network architectures to analyze such data in non-Euclidean domains, such as graphs and manifolds. Recent work  has proposed a method for constructing manifold neural networks using the spectral decomposition of the Laplace-Beltrami operator. Moreover, in this work, the authors provide a numerical scheme for implementing such neural networks when the manifold is unknown and one only has access to finitely many sample points. They show that this scheme, which relies upon building a data-driven graph, converges to the continuum limit as the number of sample points tends to infinity. Here, we build upon this result by establishing a rate of convergence that depends on the intrinsic dimension of the manifold but is independent of the ambient dimension. We also discuss how the rate of convergence depends on the depth of the network and the number of filters used in each layer.
\end{abstract}


\section{Introduction}

The emerging field of Geometric Deep Learning \cite{bronstein2017geometric,bronstein2021geometric} aims to extend the success of convolutional neural networks to more general domains such as graphs and manifolds. Though only in its infancy, this field has already achieved substantial success. For instance, Graph Neural Networks are used to power both Amazon's product recommender system \cite{Wang2022} and Google Maps \cite{derrow2021eta}. However, 
while there has been a tremendous body of research on the construction of neural networks for graphs  \cite{wu2020comprehensive}, there has been relatively little work on the development of \emph{neural networks for manifolds,} except in the case of two-dimensional surfaces \cite{masci2015geodesic}. This is in spite of the fact that manifold learning techniques \cite{tenenbaum:isomap2000,coifman:diffusionMaps2006} are often used in unsupervised algorithms for representing high-dimensional by capturing the intrinsic structure of the data. Future developments in the development of manifold neural networks could have tremendous applications in the processing of high-dimensional data arising, for example, in single-cell imaging \cite{moon2017phate,van2018recovering,amir2013visne}.

As an important step in the development of manifold neural networks, several recent works \cite{wang2021stability, wang2021stabilityrel}  have introduced CNN-like architectures that use filters based on the eigendecomposition of the Laplace-Beltrami operator and analyzed their stability properties. However, \cite{wang2021stability} and \cite{wang2021stabilityrel} assume that one knows the manifold which is not realistic for many real-world applications. Instead, one is given access to finitely many sample points $x_1,\ldots,x_n\in\mathbb{R}^D$ and makes the modeling assumption that these sample points lie upon (or near) an unknown $d$-dimensional Riemannian manifold for some $d\ll D$. In this setup, it is non-trivial to actually implement a neural network since one does not have global knowledge of the manifold. One approach, based on \textit{manifold learning} \cite{coifman:diffusionMaps2006,belkin:laplacianEigen2003,belkin2007convergence} is to construct a data-driven graph $\mathbf{G}_n$, whose vertices are the sample points $x_1,\ldots,x_n$, and use the eigenvectors and eigenvalues of graph Laplacian to approximate the eigenfunctions and eigenvalues of the Laplace-Beltrami operator. 

Recently, in \cite{wang2022convolutional}, the authors showed that if the graph $\mathbf{G}_n$ is constructed properly,  their proposed neural network will converge to a continuum limit as  the number of sample points $n$ tends to infinity. At the same time, a different line of work \cite{chew2022geometric} proposed a numerical method for implementing the manifold scattering transform \cite{perlmutter:geoScatCompactManifold2020, chew2022manifold}, a wavelet-based model of deep learning on manifolds, in the setting where one only has to sample finitely many points and proved convergence to a continuum limit. Moreover, using results from \cite{cheng2021eigen} the authors are able to establish a quantitative rate of convergence which depends on the intrinsic dimension $d$, but not on the ambient dimension $D$. 

The purpose of this paper is to apply results from \cite{chew2022geometric} and \cite{cheng2021eigen} to the manifold neural network considered in \cite{wang2022convolutional} and  establish a quantitative rate of convergence.
%\footnote{We would like to note that we are aware, based on personal correspondence, that the authors of \cite{wang2022convolutional} are independently developing a forthcoming work aiming to provide a convergence rate.  it is likely that the assumptions of their result differ from ours. (See for example Definitions 1 and 2 of \cite{wang2022singlefilter}.) }
%\footnote{We would like to note, based on personal correspondence in which we shared a preliminary version of our analysis, that the authors of \cite{wang2022convolutional} are independently developing a forthcoming work aiming to provide a convergence rate. We have not seen this work, and it is not publicly available at the present. However, based on work such as \cite{wang2022singlefilter},  it is likely that the assumptions of their result differ from ours. (See for example Definitions 1 and 2 of \cite{wang2022singlefilter}.) }
Notably, as in \cite{chew2022geometric}, our rate of convergence will depend on the intrinsic dimension $d$, but not on the ambient dimension $D$. We do not know if our rate of convergence is optimal, but it is likely that some dependence on $d$ cannot be avoided. Additionally, we note that our rate also depends on the number of layers $L$ and the number of filters $F_l$ used in each layer. As we shall discuss below, these dependencies will give insights into how to best design manifold neural networks.

\subsection{Notation}
We assume that $\mathcal{M}$ is a compact $d$-dimensional Riemannian manifold without boundary embedded in $\mathbb{R}^D$ for some $D \gg d$ such that $\text{vol}(\mathcal{M})=1$, and that the points $\{x_i\}_{i = 1}^{n}$ are uniformly sampled from $\mathcal{M}$. We will let $\mathbf{G}_n$ denote a graph with vertices $x_i$, $1\leq i \leq n$. We will let $\mathcal{L}$ denote the Laplace-Beltrami operator on $\mathcal{M}$ and let $\mathbf{L}_n$ denote the graph Laplacian on $\mathbf{G}_n$. We will let $\phi_i$ and $\lambda_i$ denote the eigenfunctions/values of the Laplace Beltrami operator $\mathcal{L}$ and let $\phi_i^n$ and $\lambda_i^n$ denote the eigenvectors/values of the graph Laplacian $\mathbf{L}_n$. We will let $\mathbf{P}_n$ be the projection operator, defined for continuous functions $f\in\mathbf{L}^2(\mathcal{M})$ by $[\mathbf{x}_n]_i=[\mathbf{P}_nf]_i=f(x_i)$. We will let $\|\cdot\|_{\mathbf{L}^2(\mathcal{M})}$ and $\|\cdot\|_{\mathbf{L}^2(\mathbf{G}_n)}$ denote the $\mathbf{L}^2$ norms on $\mathcal{M}$ and $\mathbf{G}_n.$ However, when no ambiguity exists we may simple write $\|\cdot\|$ to refer to the $\mathbf{L}^2$ norm on the appropriate space.

\section{Manifold Neural Networks}

Let $\mathcal{M}$ be a compact $d$-dimensional Riemannian manifold without boundary embedded in $\mathbb{R}^D$ for some $D \gg d$, and let $\mathcal{L}$ be the Laplace-Beltrami operator on $\mathcal{M}$. It is well-known that $\mathcal{L}$ has an orthonormal basis of eigenfunctions $\{\phi_i\}_{i=1}^\infty$ with $\mathcal{L}\phi_i=\lambda_i\phi_i$, $\lambda_i\geq 0$. This implies that for $f\in\mathbf{L}^2(\mathcal{M})$ we may write
$$f=\sum_{i=1}^\infty \widehat{f}(i) \phi_i,$$
where, for $1\leq i <\infty$, $\widehat{f}(i)$ is the \textit{generalized Fourier coefficient} defined by $\langle f,\phi_i\rangle_{\mathbf{L}^2(\mathcal{M})}$. Since we interpret the $\widehat{f}(i)$ as generalized Fourier coefficients, we will refer to a function $f$ as bandlimited if there are finitely many nonzero $\widehat{f}(i)$ as formalized in the following definition. 
\begin{definition}[Bandlimited functions] For $\kappa\geq 0,$
a function $f\in\mathbf{L}^2(\mathcal{M})$ is said to be $\kappa$-bandlimited if we have $\widehat{f}(i)=0$ for all $i>\kappa$. 
\end{definition}

Motivated by the convolution theorem in real analysis, we will define manifold convolution as  multiplication in the Fourier domain. Given a sufficiently nice function $\widehat{h}:[0,\infty)\rightarrow\mathbb{R}$, we let $h\in\mathbf{L}^2(\mathcal{M})$ be the function 
$$ h=\sum_{i=1}^\infty \widehat{h}(\lambda_i)\phi_i$$
and define a spectral convolution operator, $h(\mathcal{L})$ by 
\begin{equation}\label{eqn: specconvcontinuum} h(\mathcal{L})f=\sum_{i=1}^\infty \widehat{h}(\lambda_i) \widehat{f}(i) \phi_i.
\end{equation}Importantly, we note that since these spectral convolution operators are defined in terms of a function $\widehat{h}:[0,\infty)\rightarrow \mathbb{R}$ one may verify that the $h(\mathcal{L})$ does not depend on the choice of the orthonormal basis $\{\phi_i\}_{i=1}^\infty$. (See for example Remark 1 of \cite{chew2022geometric}.)

In \cite{wang2022convolutional}, the authors use this notion of convolution to define an $L$-layer manifold neural network with $F_\ell$ filters used in layer $\ell$. 
In this network, one is given an initial signal $f=f_0\in\mathbf{L}^2(\mathcal{M})$ and then produces a hidden representation of this signal through a multi-layer sequence of linear and non-linear maps. In particular, for $1\leq \ell\leq L$ and $1\leq p\leq F_\ell$, they define 
\begin{equation}\label{eqn: continuum MNN}f_\ell^p=\sigma\left(\sum_{q=1}^{F_{\ell-1}}h_\ell^{pq}(\mathcal{L})f_{\ell-1}^q\right), \end{equation}
where $h^{pq}_\ell(\mathcal{L})$ is a spectral convolution operator defined as in \eqref{eqn: specconvcontinuum} and $\sigma$ is a pointwise Lipschitz nonlinearity which is non-expansive in the sense that $|\sigma(a)-\sigma(b)|\leq |a-b|.$ Admissible choices of $\sigma$ include ReLu, absolute value, and many others.  
As in \cite{wang2022convolutional}, we will write the network compactly as $$\mathbf{H}=\cup_{\ell=1}^L\{h^{pq}_\ell, 1\leq q \leq F_{\ell-1}, 1\leq p \leq F_\ell\}$$ 
denote the set of all filters and let  $\boldsymbol{\Phi}(\mathbf{H},\mathcal{L},f)$ 
denote the output of their network. 
Throughout this paper, we will assume that all of the $h^{pq}_\ell$ are non-amplifying and Lipschitz as defined below.
\begin{definition}[Non-amplifying filters]
A manifold filter $h(\mathcal{L})$ is called non-amplifying  if $\|\widehat{h}\|_\infty \leq 1$.
\end{definition}
\begin{definition}[Lipschitz filters]
For $C>0$, a manifold filter $h(\mathcal{L})$ is called $C$-Lipschitz  if for all $a,b\in[0,\infty)$ we have \begin{equation*}|\widehat{h}(a)-\widehat{h}(b)| \leq C|a-b|.\end{equation*}
\end{definition}
\noindent We note that the condition that $h(\mathcal{L})$ is non-amplifying allows one to verify (using Parseval's Theorem) that $\|h(\mathcal{L})f\|_{\mathbf{L}^2(\mathcal{M})}\leq \|f\|_{\mathbf{L}^2(\mathcal{M})}$ for all $f\in\mathbf{L}^2(\mathcal{M})$ and the condition that $h$ is Lipschitz implies that $h(\mathcal{L})$ is robust to minor perturbations of the eigenvalues.
 
 As in, e.g., \cite{wang2022convolutional} and in Section 6 of \cite{chew2022geometric}, we will assume that we have a collection of $n$ points $x_1,\ldots,x_n$, which lie on $\mathcal{M}$ and are drawn i.i.d. uniformly at random with respect to the Riemannian volume form, and build a data-driven graph, $\mathbf{G}_n$, whose adjacency matrix $\mathbf{A}_n$ is defined by 
\begin{equation}\label{eqn: graph them} [\mathbf{A}_n]_{ij} = w_{ij} = \frac{1}{n} \frac{1}{t_n (4 \pi t_n)^{d/2}} \exp \left (- \frac{\|x_i - x_j\|_2^2}{4t_n} \right ),\end{equation}
where $t_n$ is a suitably chosen scale parameter. 
Given $\mathbf{A}_n$, the diagonal degree matrix $\mathbf{D}_n$ and graph Laplacian $\mathbf{L}_n$ are defined by 
\begin{equation*}
    \mathbf{L}_n = \mathbf{D}_n - \mathbf{A}_n.
\end{equation*}
It is known that the graph Laplacian has an orthonormal basis of eigenvectors $\phi_i^n$ such that $
    \mathbf{L}_n \phi_i^n = \lambda_i^n \phi_i^n,$ $\lambda_i^n\geq 0$, which allows us to write
    $$\mathbf{x}=\sum_{i=1}^n \widehat{\mathbf{x}}_i \phi_i^n, \quad \widehat{\mathbf{x}}_i=\langle \mathbf{x},\phi^n_i\rangle.$$
In order to implement a discrete network analogous to \eqref{eqn: continuum MNN}, we will need to define a projection operator $\mathbf{P}_n$ which maps a function $f$ onto its values at the sample points $x_i,$ $1\leq i\leq n$.
\begin{definition}[Projection operators]
For a continuous function $f\in\mathbf{L}^2(\mathcal{M})$, we define its projection onto $\mathbf{G}_n$ to be the vector $\mathbf{P}_nf$ whose entries are given by
$[\mathbf{P}_nf]_i=f(x_i)$. 
\end{definition}

Given this definition, one can then implement a discrete approximation of \eqref{eqn: continuum MNN} by setting $\mathbf{x}_{n,0}=\mathbf{P}_nf$
 and defining\begin{equation}\label{eqn: discretizedMNN}
\mathbf{x}_{n,\ell}^p = \sigma\left(\sum_{q=1}^{F_{\ell-1}}h^{pq}_\ell(\mathbf{L}_n)\mathbf{x}^q_{n,\ell-1}\right),
\end{equation}
where analogous to \eqref {eqn: specconvcontinuum}, $h(\mathbf{L}_n)$ is defined by 
\begin{equation*} h(\mathbf{L}_n)\mathbf{x}=\sum_{i=1}^n \widehat{h}(\lambda_i) \widehat{\mathbf{x}}_i \phi_i^n.
\end{equation*}

With these definitions, we may now state the following result which is Proposition 1 of \cite{wang2022convolutional}. It shows that this discrete approximation converges to the continuum limit under certain assumptions. 

\begin{theorem}[Proposition 1 of \cite{wang2022convolutional}]\label{thm: converge in probability}
Assume that the $x_i$ are drawn i.i.d. uniform at random from $\mathcal{M}$ and that $f$ is $\kappa$-bandlimited for some finite $\kappa$. Let $\mathbf{H}=\cup_{\ell=1}^L\{h^{pq}_\ell, 1\leq q \leq F_{\ell-1}, 1\leq p \leq F_\ell\},$ and assume that each $h\in\mathbf{H}$ is non-amplifying and $C$-Lipschitze for some finite $C>0$. Assume that $\mathbf{G}_n$ is constructed as in \eqref{eqn: graph them}, and let  $\boldsymbol{\Phi}(\mathbf{H},\mathcal{L},f)$ and $\boldsymbol{\Phi}(\mathbf{H},\mathbf{L_n},\mathbf{P}_nf)$ denote the output of the continuous and discretized   manifold neural networks defined in \eqref{eqn: continuum MNN} and \eqref{eqn: discretizedMNN}. Then, letting  $\mathbf{P_n}\boldsymbol{\Phi}(\mathbf{H},\mathcal{L},f)$ denote the projection of the output of the continuum network onto $\mathbf{G}_n$, we have  
\begin{equation*}
\lim_{n\rightarrow \infty} \|\boldsymbol{\Phi}(\mathbf{H},\mathbf{L_n},\mathbf{P}_nf)-\mathbf{P_n}\boldsymbol{\Phi}(\mathbf{H},\mathcal{L},f)\|_{\mathbf{L}^2(G_n)}=0
\end{equation*}
where the above limit is in probability.
\end{theorem}

While Theorem \ref{thm: converge in probability} guarantees convergence, it does not tell provide a quantitative rate. However, recent work by Chew et al. \cite{chew2022geometric}, provides methods for implementing the manifold scattering transform\footnote{The manifold scattering transform is a nonlinear convolutional architecture similar to that discussed here. It differs by assuming that the filters are wavelets with a certain specified form and by not featuring a summation over channels.} from finitely many sample points with a quantitative rate. 
Similar to the method discussed above, the 
method used in \cite{chew2022geometric} also relies upon building a data-driven graph $\mathbf{G}_n$ although it builds this graph in a slightly different way, defining the adjacency matrix, diagonal degree matrix, and graph Laplacian by 
\begin{align}
    [\mathbf{A}_{n}]_{i,j} &={t_n}^{-d/2}\exp\left(-\frac{\|x-x'\|^2_{2}}{t_n}\right), \nonumber \\
    [\mathbf{D}_{n}]_{i,i} &=\sum_{j=1}^n[\mathbf{A}_{n}]_{i,j}, \nonumber \\
    \mathbf{L}_n &=\frac{1}{nt_n}\left(\mathbf{D}_n-\mathbf{A}_n\right).\label{eqn: new graph}
\end{align}
\setlength{\arraycolsep}{5pt}
With the graph defined this way, they are then able to utilize the following theorem, which is a special case of  Theorem 5.4 of \cite{cheng2021eigen}.
\begin{theorem}[Theorem 5.4 of \cite{cheng2021eigen}]\label{thm: 5.4 of Chen and Wu}
Assume that the points $\{x_i\}_{i=1}^{n}$ are drawn i.i.d uniformly at random. Let $\kappa>0$ be fixed. Assume that the first $\kappa$ eigenvalues of the (true) Laplace-Beltrami operator, $\mathcal{L},$ $\lambda_1,\ldots,\lambda_{\kappa}$, all have single multiplicity. Construct $\mathbf{G}_n$ as in \eqref{eqn: new graph}, and let $\phi_i^n$ and $\lambda_i^{n}$  be the eigenvectors and eigenvalues of the data-driven $\mathbf{L}_n$.   Assume that $t_n\rightarrow 0$ and $n\rightarrow\infty$ at a rate where $t_n \sim n^{-2/(d+6)}$. Then, with probability at least $1-\mathcal{O}\left(\frac{1}{n^9}\right)$,
there exist scalars $\alpha_k$ with
\begin{equation}\label{eqn: alpha error} |\alpha_k|=1+o(1)\end{equation} such that for all $0\leq i\leq \kappa$
\begin{equation*}
    |\lambda_i-\lambda^{n}_i|=\mathcal{O}\left(n^{-\frac{2}{d+6}}\right)
\end{equation*}
    and
\begin{equation*}
    \|\mathbf{P}_n\phi_i-\alpha_i\phi_i^n\|_2=\mathcal{O}\left(n^{-\frac{2}{d+6}}\sqrt{\log n}\right),
\end{equation*}
 where the constants implied by the big-$\mathcal{O}$ notation depend on $\kappa$ and the geometry of $\mathcal{M}$.
\end{theorem}

\begin{remark}\label{rem: evec rate}
As noted earlier, spectral filters of the form \eqref{eqn: specconvcontinuum} do not depend on the choice of eigenbasis. Therefore, when applying Theorem \ref{thm: 5.4 of Chen and Wu} to the manifold neural networks considered here we may assume that the $\alpha_i>0$ (since otherwise we change the basis by replacing $\phi_i$ with $-\phi_i$.) 

Therefore, by imitating the proof of Theorem 10 of \cite{chew2022geometric}, with probability at least $1 - \mathcal{O}(\frac{1}{n^9})$, we may see
\begin{equation*}
    \|\mathbf{P}_n\phi_i-\mathbf{\phi}_i^{n}\|_2 = \mathcal{O}\left(n^{-\frac{2}{d+6}}\sqrt{\log n}\right).
\end{equation*}
\end{remark}

\noindent The authors of \cite{chew2022geometric}  also make frequent use of the following Lemma which is a consequence of Hoeffding's inequality.
\begin{lemma}[Lemma 5 of \cite{chew2022geometric}]\label{lem: apply Hoeffding}Assume that the points $\{x_i\}_{i=1}^{n}$ are drawn i.i.d uniformly at random, and
let $f,g\in\mathcal{C}(\mathcal{M})$. Then, with probability at least $1-\frac{2}{n^9}$ we have 
%then with probability at least $1-2\exp\left(\frac{-N\eta^2}{\|fg\|_\infty}\right)$ we have 
%\begin{equation*}
 %   |\langle \rho f,\rho g \rangle_2 - \langle f,g\rangle_{\mathbf{L}^2(\mathcal{X})}| \leq \eta.
%\end{equation*}
%In particular, if   $\eta = \sqrt{\frac{\log(N)\|fg\|_\infty}{N}}$, we have 
\begin{equation*}
    |\langle  f, g \rangle_{\mathbf{G}_n} - \langle f,g\rangle_{\mathbf{L}^2(\mathcal{M})}| \leq  \sqrt{\frac{18 \log n}{n}}\|fg\|_\infty.%\sqrt{\frac{9\log N}{N}}\sqrt{\|fg\|_\infty}. 
\end{equation*}
%with probability at least $1-\frac{2}{N}$.
\end{lemma}

Given Theorem \ref{thm: 5.4 of Chen and Wu} and Lemma \ref{lem: apply Hoeffding}, the authors of \cite{chew2022geometric} then show that the manifold scattering transform converges at a quantifiable rate informally summarized below in the following theorem. (See Theorems 13 and 14 \cite{chew2022geometric} for a precise statement).
\begin{theorem}[Informal]
Under the assumptions of Theorem \ref{thm: 5.4 of Chen and Wu}, the manifold scattering transform converges at rate $\mathcal{O}\left(n^{-\frac{2}{d+6}}\sqrt{\log n}\right)$.
\end{theorem}
The main result of this paper, discussed in the following section, establishes a similar convergence result for manifold neural networks. 
\section{Convergence Rates for Manifold Neural Networks}

The following theorem is our main result. It builds upon Theorem \ref{thm: converge in probability} by providing a quantitative rate of convergence. Its proof is based on combining Theorem \ref{thm: 5.4 of Chen and Wu} with the techniques used to prove Theorem \ref{thm: converge in probability}. Full details are provided in the supplementary material. 
\begin{theorem}\label{thm: mnn with rate}
Assume that the $x_i$ are drawn i.i.d. uniformly at random from $\mathcal{M}$ and that $f$ is $\kappa$ bandlimited. Assume that each of the filters $h\in\mathbf{H}$ is non-amplifying and $C$-Lipschitz, and let $\tilde{C} = \max \{C, 1\}$.
Assume that $\mathbf{G}_n$ is constructed as in \eqref{eqn: new graph}, let  $\boldsymbol{\Phi}(\mathbf{H},\mathcal{L},f)$ and $\boldsymbol{\Phi}(\mathbf{H},\mathbf{L_n},\mathbf{P}_nf)$ denote the output of the continuous and discretized   manifold neural networks defined in \eqref{eqn: continuum MNN} and \eqref{eqn: discretizedMNN} and assume that $t_n\rightarrow 0$ and $n\rightarrow\infty$ at a rate where $t_n \sim n^{-2/(d+6)}$. Then with probability at least
$1 - \mathcal{O}(\frac{1}{n^9})$, we have that
\begin{align*}
 &\|\boldsymbol{\Phi}(\mathbf{H},\mathbf{L_n},\mathbf{P}_nf)-\mathbf{P_n}\boldsymbol{\Phi}(\mathbf{H},\mathcal{L},f)\|_{\mathbf{L}^2(G_n)} \\
 &\leq\tilde{C}\mathcal{O}\left(\frac{\sqrt{\log n}}{n^{2/(d+6)}}\right)\max_{\ell,q} \|f^q_\ell\|\sum_{k=1}^{L} \prod_{j=L-k}^{L} F_j\\
 &+\tilde{C}\mathcal{O}\left ( \frac{(\log n)^{1/4}}{n^{1/4 + 2/(d+6)}} \right)\max_{\ell,q} \|f^q_\ell\|_\infty \sum_{k=1}^{L} \prod_{j=L-k}^{L} F_j
\end{align*}
when $d \geq 2$. In the case when $d = 1$, we instead have 
\begin{align*}
 &\|\boldsymbol{\Phi}(\mathbf{H},\mathbf{L_n},\mathbf{P}_nf)-\mathbf{P_n}\boldsymbol{\Phi}(\mathbf{H},\mathcal{L},f)\|_{\mathbf{L}^2(G_n)}\\
 &\leq \tilde{C}\mathcal{O}\left( \frac{\sqrt{\log n}}{n^{2/7}}\right )\max_{\ell,q} \|f^q_\ell\| \sum_{k=1}^{L} \prod_{j=L-k}^{L} F_j\\
 &+\tilde{C} \mathcal{O}\left ( \frac{\sqrt{\log n}}{n^{1/2}}\right)\max_{\ell,q} \|f^q_\ell\|_\infty \sum_{k=1}^{L} \prod_{j=L-k}^{L} F_j.
\end{align*}
\end{theorem}

In order to interpret Theorem \ref{thm: mnn with rate}, we may make the following observations on the rate of convergence. \begin{itemize}
    \item Our bounds depend on the intrinsic dimension $d$ but not on the ambient dimension $D$, and the rate of convergence with respect to $n$ is the same as that of analogous results in \cite{chew2022geometric}. We also note that our rate is slightly faster than the rate of $\mathcal{O}(n^{-1/(2d+8)})$ which was established for a single filter (satisfying certain assumptions) in Theorem 3 of \cite{wang2022singlefilter}.
    \item The rate of convergence depends on the $L^2$ and $L^\infty$ norms of the filters (although for large $n$ the $L^2$ term dominates). Therefore, when training manifold neural networks, one might want to employ regularization when training the filters.
    \item The bound depends linearly on the Lipschitz constant of the filters. Again, this suggests that one might want to employ regularization.
    \item Networks with more filters (unsurprisingly) require more training data. More specifically, if one uses an $L$ layer network with $F_\ell=F$ filters in each layer, then our bound depends on $F$ via 
    $$\sum_{k=1}^{L} \prod_{j=L-k}^{L} F_j   \approx F^{L+1}.$$ 
    Therefore, very deep networks may require large amounts of training data and so it might be better to use a fairly small number of layers  unless one has truly massive amounts of data. Interestingly, this in some sense parallels a well-known phenomenon in graph neural networks where performance typically either levels or deteriorates if more than two or three layers are used (although this is likely for a different reason). This is in contrast to Euclidean CNNs where optimal performance is usually obtained with much deeper networks.
\end{itemize}

\section{Numerical Experiments}
\begin{figure}[h]
    \centering
\includegraphics[width=\linewidth]{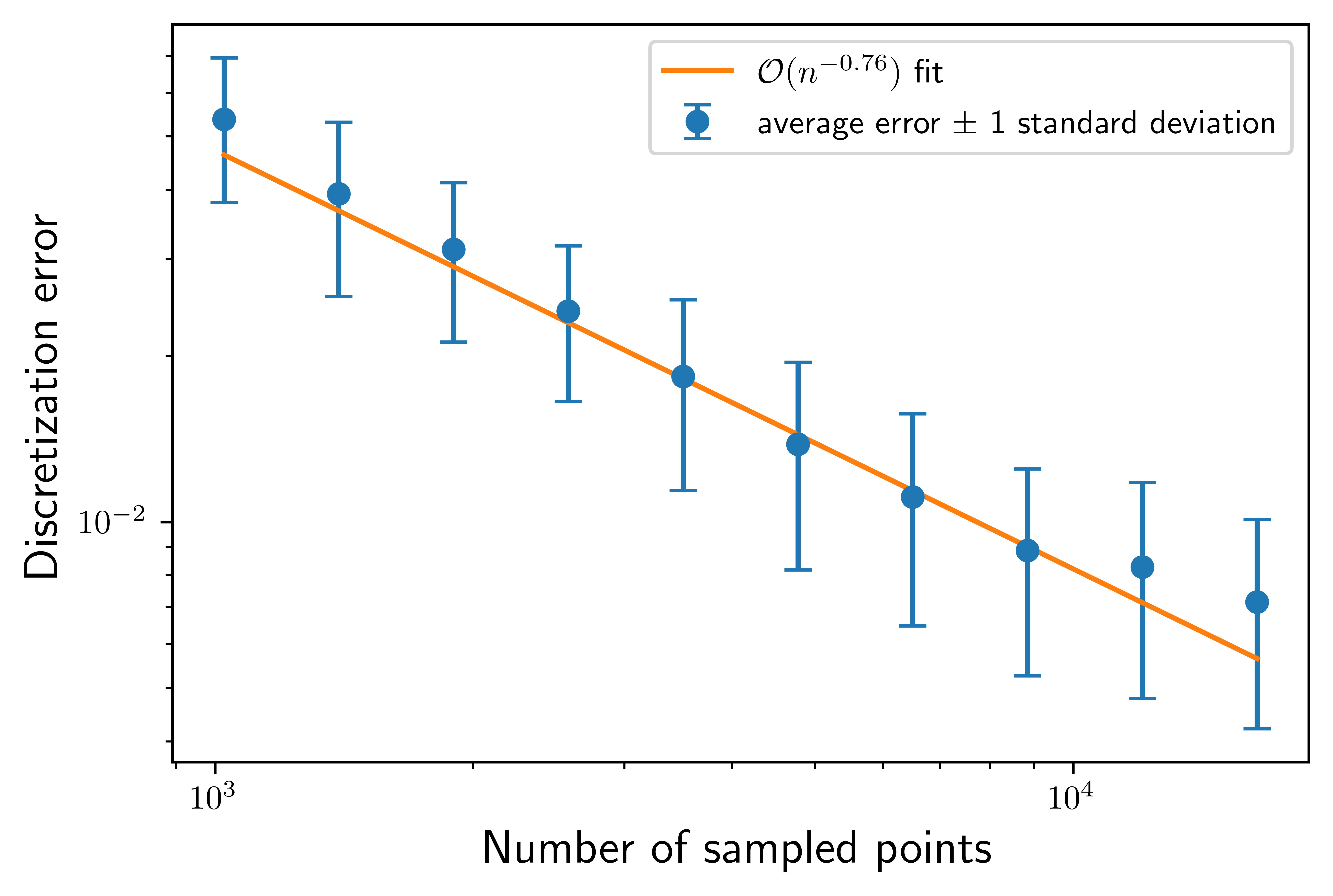}
    \caption{Errors and log-log fit over 10 logarithmically spaced values of $n$ from $2^{10}$ to $2^{14}$ points. Both axes are logarithmically scaled.}
    \label{fig:convergence rate}
\end{figure}

We investigate the convergence rate of the proposed MNN with experiments on the two-dimensional sphere, which enables exact evaluation of the continuous MNN along with the discretized MNN. We uniformly sample up to $2^{14}$ points on the sphere and construct a signal $f$ by
\begin{equation*}
    f = \sum_{i=1}^9 \alpha_i \phi_i
\end{equation*}
where $\alpha_i \in \mathbb{R}$ and $\phi_i$ are the spherical harmonics, which are the eigenfunctions of the Laplace-Beltrami operator on the sphere. Hence, $\widehat{f}(i) = \alpha_i$ and thus we can exactly compute $h(\mathcal{L})f$. We use a single layer with one filter defined by $\widehat{h}(\lambda) = e^{-\lambda}$ and use the absolute value as the nonlinearity $\sigma$. In Figure \ref{fig:convergence rate}, we plot $\|\boldsymbol{\Phi}(\mathbf{H},\mathbf{L_n},\mathbf{P}_nf)-\mathbf{P_n}\boldsymbol{\Phi}(\mathbf{H},\mathcal{L},f)\|_{\mathbf{L}^2(G_n)}$ against the number of sampled points $N$, averaged over 100 trials. We observe a convergence rate of about $\mathcal{O}(n^{-0.76})$ which is faster than the rate guaranteed by our Theorem in the case where $d=2$,  $\mathcal{O}(n^{-1/4})$. This suggests that it may be possible to sharpen our rate of convergence in future work, at least in cases such as the sphere where we have additional information about the structure of the eigenfunctions.

\section{Conclusions and Future Work}
We have established a quantitive rate of convergence for the manifold neural network introduced in \cite{wang2021stability, wang2021stabilityrel}. Notably, our rate of convergence depends on the depth of the network and the number of filters used in each layer suggesting that truly deep networks may require massive amounts of training data. This raises an important question for future inquiry. Can one design a manifold neural network whose sampling requirements scale at most linearly with respect to the depth of the network and the number of filters used per layer?

Additionally, we note that both Theorem \ref{thm: converge in probability} (restated from \cite{wang2022convolutional}) and Theorem \ref{thm: mnn with rate} assume that the sample points $x_i$ lie exactly on the manifold $\mathcal{M}$ and are sampled i.i.d. uniformly with respect to the Riemannian volume form. We view removing these assumptions as an important line of future inquiry. One possible solution for addressing the first issue is to use techniques from noise-robust manifold learning (see \cite{landa2022robust} and the references therein) in which one assumes the data has the form $y_i=x_i+\eta_i$ where $x_i\in\mathcal{M}$ and $\eta_i$ is noise. Similarly, for the second challenge, one could consider methods used in works such as \cite{coifman:diffusionMaps2006,dunson2021spectral, Little2022Balancing} where the data is assumed to be sampled from a non-uniform density. However, adapting these results to our setting is non-trivial because most of these works establish pointwise convergence guarantees whereas our proof techniques require convergence in $\mathbf{L}^2$.

\section*{Acknowledgments}

% The preferred spelling of the word ``acknowledgment'' in America is without 
% an ``e'' after the ``g''. Avoid the stilted expression ``one of us (R. B. 
% G.) thanks $\ldots$''. Instead, try ``R. B. G. thanks$\ldots$''. Put sponsor 
% acknowledgments in the unnumbered footnote on the first page.
Joyce Chew was supported by NSF DGE 2034835. Deanna Needell was supported by NSF DMS 2011140 and NSF DMS 2108479.
\bibliographystyle{IEEEtran}
\bibliography{IEEEabrv,main}

\newpage
\appendices
\section{Proof of \Cref{thm: mnn with rate}}\label{sec: proof}
\begin{proof}[The Proof of Theorem \ref{thm: mnn with rate}]
      
We first note that
    \begin{align*}
        \|\boldsymbol{\Phi}(\mathbf{H},\mathbf{L_n},\mathbf{P}_nf)-\mathbf{P_n}\boldsymbol{\Phi}(\mathbf{H},\mathcal{L},f)\| &\leq \sum_{q=1}^{F_L}\|\mathbf{x}_{n,L}^q-\mathbf{P}_nf_L^q\|.
    \end{align*}
    Next, we observe that for any $1\leq \ell\leq L$, $1\leq p\leq F_\ell,$ we have
    \begin{align}
&\|\mathbf{x}_{n,\ell}^p-\mathbf{P}_nf_\ell^p\|\nonumber\\
\leq& \sum_{q=1}^{F_{\ell-1}}\|h^{pq}_\ell(\mathbf{L}_n)\mathbf{x}^q_{n,\ell-1}-\mathbf{P}_nh^{pq}_\ell(\mathcal{L})f^q_{\ell-1}\|\nonumber\\
\leq& \sum_{q=1}^{F_{\ell-1}}\|h^{pq}_\ell(\mathbf{L}_n)\mathbf{x}^q_{n,\ell-1}-h^{pq}_\ell(\mathbf{L}_n)\mathbf{P}_nf^q_{\ell-1}\| \nonumber\\
&+\sum_{q=1}^{F_{\ell-1}}\|h^{pq}_\ell(\mathbf{L}_n)\mathbf{P}_nf^q_{\ell-1}-\mathbf{P}_nh^{pq}_\ell(\mathcal{L})f^q_{\ell-1}\|.\label{eqn: setup recursion}
    \end{align}
Since  $h^{pq}_\ell$ is non-amplifying, the matrix $h^{pq}_\ell(\mathbf{L}_n)$ has operator norm at most one on $\mathbf{L}^2(\mathbf{G}_n)$. Therefore, we may bound the first term from \eqref{eqn: setup recursion} by
\begin{align}
\sum_{q=1}^{F_{\ell-1}}&\|h^{pq}_\ell(\mathbf{L}_n)\mathbf{x}^q_{n,\ell-1}-h^{pq}_\ell(\mathbf{L}_n)\mathbf{P}_nf^q_{\ell-1}\| \nonumber\\
&\leq F_{\ell-1}\max_{1\leq q \leq F_{L-1}}\|\mathbf{x}^q_{n,\ell-1}-\mathbf{P}_nf^q_{\ell-1}\|.\label{eqn: easy recursion term}
\end{align}
For notational clarity, we will now temporarily drop the feature index $q$ and the layer index $l$. To bound the second term, we note that:
\begin{align}
&\|h(\mathbf{L}_n)\mathbf{P}_nf-\mathbf{P}_nh(\mathcal{L})f\|\nonumber\\
= &\left\| \sum_{i=1}^\kappa \widehat{h}(\lambda_i^n)\langle \mathbf{P}_nf,\phi_i^n\rangle_{\mathbf{G}_n}\phi_i^n - \sum_{i=1}^\kappa \widehat{h}(\lambda_i)\langle f,\phi_i\rangle_\mathcal{M}\mathbf{P}_n\phi_i\right\|\nonumber\\
\leq &\left\| \sum_{i=1}^\kappa (\widehat{h}(\lambda_i^n)-\widehat{h}(\lambda_i))\langle \mathbf{P}_nf,\phi_i^n\rangle_{\mathbf{G}_n}\phi_i^n\right\|\label{eqn: different eigenvalues}\\
&+\left\|\sum_{i=1}^\kappa \widehat{h}(\lambda_i)(\langle \mathbf{P}_nf,\phi_i^n\rangle_{\mathbf{G}_n}\phi_i^n- \langle f,\phi_i\rangle_\mathcal{M}\mathbf{P}_n\phi_i)\right\|.\label{eqn: different eigenvectors}
\end{align}

To bound the term from \eqref{eqn: different eigenvalues}, we note that by the triangle inequality and the Cauchy-Schwarz inequality we have
\begin{align*}
    &\left\| \sum_{i=1}^\kappa (\widehat{h}(\lambda_i^n) - \widehat{h}(\lambda_i))\langle \mathbf{P}_nf,\phi_i^n\rangle_{\mathbf{G}_n}\phi_i^n\right\|\\
    &\qquad\qquad\leq \max_{1\leq i \leq \kappa} |\widehat{h}(\lambda_i^n)- \widehat{h}(\lambda_i)| \sum_{i=1}^\kappa \|\mathbf{P}_n f\| \|\phi_i^n\|^2.
\end{align*}
By \Cref{rem: evec rate} and the assumption that all of the $h^{pq}_\ell$ are  $C$-Lipschitz, we have 
\begin{equation*}
    \max_{1\leq i \leq \kappa} |\widehat{h}(\lambda_i^n)- \widehat{h}(\lambda_i)|\leq C\mathcal{O}\left(n^{-\frac{2}{d+6}}\sqrt{\log n}\right),
\end{equation*}
and by Lemma \ref{lem: apply Hoeffding}, we have 
\begin{equation*}
\|\mathbf{P}_n f\|^2 \leq \|f\|_2^2 + \sqrt{\frac{18\log n}{n}} \|f\|_\infty^2.
\end{equation*}
Therefore, using fact that  $\sqrt{a^2+b^2}\leq |a|+|b|$ for all $a,b\in\mathbb{R}$, we obtain
\begin{align}
    &\left\| \sum_{i=1}^\kappa (\widehat{h}(\lambda_i^n) - \widehat{h}(\lambda_i))\langle \mathbf{P}_nf,\phi_i^n\rangle_{\mathbf{G}_n}\phi_i^n\right\|\nonumber\\
    &\leq C\kappa \mathcal{O}\left ( \frac{\sqrt{\log{n}}}{n^{2/(d+6)}}\right )\sqrt{\|f\|_2^2 + \sqrt{\frac{18\log n}{n}} \|f\|_\infty^2} \nonumber\\ 
    &\leq C\kappa \mathcal{O}\left ( \frac{\sqrt{\log{n}}}{n^{2/(d+6)}}\right )\left(\|f\|_2 + \mathcal{O}\left ( \left (\frac{\log n}{n} \right )^{1/4}\right) \|f\|_\infty\right).\label{eqn: eval diff}
    \end{align}
Now, turning our attention to the terms from \eqref{eqn: different eigenvectors}, we have 
\begin{align}
    &\left\|\sum_{i=1}^\kappa \widehat{h}(\lambda_i)(\langle \mathbf{P}_nf,\phi_i^n\rangle_{\mathbf{G}_n}\phi_i^n- \langle f,\phi_i\rangle_\mathcal{M}\mathbf{P}_n\phi_i)\right\|\nonumber\\
    &\leq \left\|\sum_{i=1}^\kappa \widehat{h}(\lambda_i)\langle \mathbf{P}_nf,\phi_i^n\rangle_{\mathbf{G}_n}(\phi_i^n-\mathbf{P}_n\phi_i)\right\|\label{eqn: use eigenvector rate}\\
    &\quad+\left\|\sum_{i=1}^\kappa \widehat{h}(\lambda_i)(\langle \mathbf{P}_nf,\phi_i^n\rangle_{\mathbf{G}_n}- \langle f,\phi_i\rangle_\mathcal{M})\mathbf{P}_n\phi_i\right\|\label{eqn: use Hoeffding this time}.
\end{align}
By \Cref{rem: evec rate}, we have $\|\phi_i^n-\mathbf{P}_n\phi_i\|=\mathcal{O}\left(n^{-\frac{2}{d+6}}\sqrt{\log n}\right)$. Therefore, since the frequency responses of $\mathbf{H}$ are non-amplifying, the term from \eqref{eqn: use eigenvector rate} can be bounded by 
\begin{align}
    &\left\|\sum_{i=1}^\kappa \widehat{h}(\lambda_i)\langle \mathbf{P}_nf,\phi_i^n\rangle_{\mathbf{G}_n}(\phi_i^n-\mathbf{P}_n\phi_i)\right\|\nonumber\\
    &\leq \kappa \max_{1\leq i\leq \kappa} |\langle \mathbf{P}_nf,\phi_i^n\rangle_{\mathbf{G}_n}|\|\phi_i^n-\mathbf{P}_n\phi_i\|\nonumber\\
    &\leq \kappa\mathcal{O}\left( \frac{\sqrt{\log n}}{n^{2/(d+6)}}\right )\|\mathbf{P}_nf\|\nonumber\\
    &\leq \kappa\mathcal{O}\left( \frac{\sqrt{\log n}}{n^{2/(d+6)}}\right )\left (\|f\|_2+\mathcal{O}\left ( \left (\frac{\log n}{n} \right )^{1/4}\right)\|f\|_\infty\right ),\label{eqn: evec diff}
\end{align}
where the final inequality again follows by using Lemma \ref{lem: apply Hoeffding}.
Meanwhile, the term from \eqref{eqn: use Hoeffding this time} can be bounded by 
\begin{align*}
    &\left\|\sum_{i=1}^\kappa \widehat{h}(\lambda_i)(\langle \mathbf{P}_nf,\phi_i^n\rangle_{\mathbf{G}_n}- \langle f,\phi_i\rangle_\mathcal{M})\mathbf{P}_n\phi_i\right\|\\
    &\leq\sum_{i=1}^\kappa |\widehat{h}(\lambda_i)| |\langle \langle \mathbf{P}_nf,\phi_i^n\rangle_{\mathbf{G}_n}- \langle f,\phi_i\rangle_\mathcal{M}|\|\mathbf{P}_n\phi_i\|\\
    &\leq\sum_{i=1}^\kappa |\langle \mathbf{P}_nf,\phi_i^n\rangle_{\mathbf{G}_n}- \langle f,\phi_i\rangle_\mathcal{M}|\|\mathbf{P}_n\phi_i\|\\
    &\leq\sum_{i=1}^\kappa |\langle \mathbf{P}_nf,\phi_i^n\rangle_{\mathbf{G}_n}-\langle \mathbf{P}_nf,\mathbf{P}_n\phi_i\rangle_{\mathbf{G}_n}|\|\mathbf{P}_n\phi_i\|\\
    &\qquad+\sum_{i = 1}^\kappa |\langle \mathbf{P}_nf,\mathbf{P}_n\phi_i\rangle_{\mathbf{G}_n}- \langle f,\phi_i\rangle_\mathcal{M}|\|\mathbf{P}_n\phi_i\|.
\end{align*}
 By the Cauchy-Schwarz inequality, \Cref{rem: evec rate}, and Lemma \ref{lem: apply Hoeffding}, we have 
\begin{align*}
    &|\langle \mathbf{P}_nf,\phi_i^n\rangle-\langle \mathbf{P}_nf,\mathbf{P}_n\phi_i\rangle| \\
    &\leq \| \mathbf{P}_nf\| \|\phi_i^n-\mathbf{P}_n\phi_i\|\\
    &\leq \mathcal{O}\left( \frac{\sqrt{\log n}}{n^{-2/d+6}}\right)\left(\|f\|_2+\mathcal{O}\left ( \left (\frac{\log n}{n} \right )^{1/4}\right)\|f\|_\infty\right).
\end{align*}
Again using Lemma \ref{lem: apply Hoeffding}, we have 
\begin{align*}
    |\langle \mathbf{P}_nf,\mathbf{P}_n\phi_i\rangle- \langle f,\phi_i\rangle| \leq \mathcal{O}\left(\left(\frac{\log n}{n}\right)^{1/2}\right)\|f\|_\infty\|\phi_i\|_\infty
\end{align*}
and also that 
$$\|\mathbf{P}_n\phi_i\|\leq 1+\mathcal{O}\left ( \left (\frac{\log n}{n} \right )^{1/4}\right)\|\phi_i\|_\infty.$$
It is known (see Appendix L of \cite{chew2022geometric} and the references there) that $\|\phi_i\|_\infty \leq C_\mathcal{M} i^{(d-1)/2d}\leq C_\mathcal{M}i^{1/2}$. Therefore, for all $i\leq \kappa$ we have 
\begin{align*}
    |\langle \mathbf{P}_nf,\mathbf{P}_n\phi_i\rangle- \langle f,\phi_i\rangle| &\leq \mathcal{O}\left(\left(\frac{\log n}{n}\right)^{1/2}\right)\kappa^{1/2}\|f\|_\infty \\
    &=\mathcal{O}\left(\left(\frac{\log n}{n}\right)^{1/2}\right)\|f\|_\infty,
\end{align*}
(where in the second inequality we used the fact that the constants implied by big-$\mathcal{O}$ notation depend on $\kappa$). Similarly, we also have
\begin{align*}
\|\mathbf{P}_n\phi_i\|&\leq 1+\mathcal{O}\left ( \left (\frac{\log n}{n} \right )^{1/4}\right) \kappa^{1/2}\\
&= 1+\mathcal{O}\left ( \left (\frac{\log n}{n} \right )^{1/4}\right)\\
&=\mathcal{O}(1).
\end{align*}
Therefore, when $d \geq 2$, we have
\begin{align}
    &\left\|\sum_{i=1}^\kappa \widehat{h}(\lambda_i)(\langle \mathbf{P}_nf,\phi_i^n\rangle_{\mathbf{G}_n} -\langle f_{\ell-1}^q,\phi_i\rangle_\mathcal{M})\mathbf{P}_n\phi_i\right\| \nonumber\\
    \leq&\sum_{i=1}^\kappa |\langle \mathbf{P}_nf,\phi_i^n\rangle-\langle \mathbf{P}_nf,\mathbf{P}_n\phi_i\rangle|\|\mathbf{P}_n\phi_i\| \nonumber \\
    &\qquad+\sum_{i=1}^\kappa|\langle \mathbf{P}_nf,\mathbf{P}_n\phi_i\rangle- \langle f,\phi_i\rangle|\|\mathbf{P}_n\phi_i\|\nonumber\\
    \leq&\sum_{i=1}^\kappa \mathcal{O}\left(\frac{\sqrt{\log n}}{n^{2/(d+6)}}\right)\left(\|f\|_2+\mathcal{O}\left ( \left (\frac{\log n}{n} \right )^{1/4}\right)\|f\|_\infty\right )\nonumber \\
    &\qquad+\sum_{i=1}^\kappa\mathcal{O}\left(\left(\frac{\log n}{n}\right)^{1/2}\right) \|f\|_\infty \nonumber\\    
    \leq&\kappa \mathcal{O}\left(\frac{\sqrt{\log n}}{n^{2/(d+6)}}\right)\left(\|f\|_2+\mathcal{O}\left ( \left (\frac{\log n}{n} \right )^{1/4}\right)\|f\|_\infty\right ) \nonumber\\
    &\qquad+\kappa\mathcal{O}\left ( \left (\frac{\log n}{n}\right)^{1/2}\right )\|f\|_\infty\label{eqn: any d} \\
    \leq&\kappa \left(\mathcal{O}\left(\frac{\sqrt{\log n}}{n^{2/(d+6)}}\right)\left(\|f\|_2+\mathcal{O}\left ( \left (\frac{\log n}{n} \right )^{1/4}\right)\|f\|_\infty\right ) \right)\label{eqn: inner pdt diff}
    % &\leq\kappa \left(\mathcal{O}\left(n^{-\frac{2}{d+6}}\right)\left(\|f_{\ell-1}\|_2+\mathcal{O}\left ( \left (\frac{\log n}{n} \right )^{1/4}\right)\|f_{\ell-1}\|_\infty\right ) \right)
\end{align}
where in the last line we used the fact that $d \geq 2$.

Therefore, combining Equations \eqref{eqn: eval diff} through \eqref{eqn: inner pdt diff} and reinstating the feature and layer indices, we have 
\begin{align*}
&\|h^{pq}_\ell(\mathbf{L}_n)\mathbf{P}_nf^q_{\ell-1}-\mathbf{P}_nh^{pq}_\ell(\mathcal{L})f^q_{\ell-1}\|\nonumber\\
\leq &\left\| \sum_{i=1}^\kappa (\widehat{h}_\ell^{pq}(\lambda_i^n) - \widehat{h}_\ell^{pq}(\lambda_i))\langle \mathbf{P}_nf^q_{\ell-1},\phi_i^n\rangle_{\mathbf{G}_n}\phi_i^n\right\|\\
&+\left\|\sum_{i=1}^\kappa \widehat{h}_\ell^{pq}(\lambda_i)(\langle \mathbf{P}_nf^q_{\ell-1},\phi_i^n\rangle_{\mathbf{G}_n}\phi_i^n-\langle f^q_{\ell-1},\phi_i\rangle_\mathcal{M}\mathbf{P}_n\phi_i)\right\|\\
\leq& C\kappa \mathcal{O}\left ( \frac{\sqrt{\log n}}{n^{2/(d+6)}}\right )\left(\|f^q_{\ell-1}\|_2 + \mathcal{O}\left ( \left (\frac{\log n}{n} \right )^{1/4}\right) \|f^q_{\ell-1}\|_\infty\right)\\
&+\kappa\mathcal{O}\left( \frac{\sqrt{\log n}}{n^{2/(d+6)}}\right )\left (\|f^q_{\ell-1}\|_2+\mathcal{O}\left ( \left (\frac{\log n}{n} \right )^{1/4}\right)\|f^q_{\ell-1}\|_\infty\right )\\
&+\kappa \mathcal{O}\left(\frac{\sqrt{\log n}}{n^{2/(d+6)}}\right)\left(\|f^q_{\ell-1}\|_2+\mathcal{O}\left ( \left (\frac{\log n}{n} \right )^{1/4}\right)\|f^q_{\ell-1}\|_\infty\right )\\
\leq& \tilde{C}\mathcal{O}\left(\frac{\sqrt{\log n}}{n^{2/(d+6)}}\right)\left(\|f^q_{\ell-1}\|_2+\mathcal{O}\left ( \left (\frac{\log n}{n} \right )^{1/4}\right)\|f^q_{\ell-1}\|_\infty\right ),  
\end{align*}
where $\tilde{C} = \max\{C, 1\}$ and in the final line we have absorbed $\kappa$ into the implied constant.

When $d = 1$, we repeat the same string of inequalities up to \cref{eqn: any d} and obtain
\begin{align}
    &\left\|\sum_{i=1}^\kappa \widehat{h}(\lambda_i)(\langle \mathbf{P}_nf,\phi_i^n\rangle_{\mathbf{G}_n}\mathbf{P}_n\phi_i -\langle f,\phi_i\rangle_\mathcal{M})\mathbf{P}_n\phi_i\right\| \nonumber\\
    \leq&\kappa \mathcal{O}\left(\frac{\sqrt{\log n}}{n^{2/7}}\right)\left(\|f^q_{\ell-1}\|_2+\mathcal{O}\left ( \left (\frac{\log n}{n} \right )^{1/4}\right)\|f^q_{\ell-1}\|_\infty\right ) \nonumber\\
    &\qquad+\kappa\mathcal{O}\left ( \left (\frac{\log n}{n}\right)^{1/2}\right )\|f^q_{\ell-1}\|_\infty \nonumber \\
    \leq& \kappa \left ( \mathcal{O}\left(\frac{\sqrt{\log n}}{n^{2/7}}\right) \|f^q_{\ell-1}\|_2 + \mathcal{O}\left ( \frac{\sqrt{\log n}}{n^{1/2}}\right )\|f^q_{\ell-1}\|_\infty \right )\label{eqn: inner pdt diff d1}
\end{align}

Then, combining \cref{eqn: eval diff,eqn: evec diff,eqn: inner pdt diff d1}, and again absorbing $\kappa$ into the implied constant we obtain

\begin{align*}
&\|h^{pq}_\ell(\mathbf{L}_n)\mathbf{P}_nf^q_{\ell-1}-\mathbf{P}_nh^{pq}_\ell(\mathcal{L})f^q_{\ell-1}\|\nonumber\\
\leq &\left\| \sum_{i=1}^\kappa (\widehat{h}_\ell^{pq}(\lambda_i^n)-\widehat{h}_\ell^{pq}(\lambda_i))\langle \mathbf{P}_nf^q_{\ell-1},\phi_i^n\rangle_{\mathbf{G}_n}\phi_i^n\right\|\\
&+\left\|\sum_{i=1}^\kappa \widehat{h}_\ell^{pq}(\lambda_i)(\langle \mathbf{P}_nf^q_{\ell-1},\phi_i^n\rangle_{\mathbf{G}_n}\phi_i^n-\langle f^q_{\ell-1},\phi_i\rangle_\mathcal{M})\mathbf{P}_n\phi_i\right\|\\
\leq& C\kappa \mathcal{O}\left ( \frac{\sqrt{\log n}}{n^{2/7}}\right )\left(\|f^q_{\ell-1}\|_2 + \mathcal{O}\left ( \left (\frac{\log n}{n} \right )^{1/4}\right) \|f^q_{\ell-1}\|_\infty\right)\\
&+\kappa\mathcal{O}\left( \frac{\sqrt{\log n}}{n^{2/7}}\right )\left (\|f^q_{\ell-1}\|_2+\mathcal{O}\left ( \left (\frac{\log n}{n} \right )^{1/4}\right)\|f^q_{\ell-1}\|_\infty\right )\\
&+\kappa \left ( \mathcal{O}\left(\frac{\sqrt{\log n}}{n^{2/7}}\right) \|f^q_{\ell-1}\|_2 + \mathcal{O}\left ( \frac{\sqrt{\log n}}{n^{1/2}}\right )\|f^q_{\ell-1}\|_\infty \right)\\
\leq& \tilde{C}\left (\mathcal{O}\left( \frac{\sqrt{\log n}}{n^{2/7}}\right )\|f^q_{\ell-1}\|_2+\mathcal{O}\left ( \frac{\sqrt{\log n}}{n^{1/2}}\right )\|f^q_{\ell-1}\|_\infty\right).  
\end{align*}

Thus by \eqref{eqn: setup recursion} and \eqref{eqn: easy recursion term}, we have derived the relationship that, when $d \geq 2$, 
\begin{align*}
    \|\mathbf{x}_{n,\ell}^p-\mathbf{P}_nf_\ell^p\| &\leq F_{\ell-1}\max_{1\leq q \leq F_{L-1}}\|\mathbf{x}^q_{n,\ell-1}-\mathbf{P}_nf^q_{\ell-1}\| \\
    % &\quad+ \tilde{C}\mathcal{O}\left(\frac{\sqrt{\log n}}{n^{2/(d+6)}}\right)\left(\|f^q_{\ell-1}\|_2+\mathcal{O}\left ( \left (\frac{\log n}{n} \right )^{1/4}\right)\|f^q_{\ell-1}\|_\infty\right )
    &\quad+ F_{\ell -1}\tilde{C}\mathcal{O}\left(\frac{\sqrt{\log n}}{n^{2/(d+6)}}\right)\|f^q_{\ell-1}\|_2 \\
    &\quad+F_{\ell -1}\tilde{C}\mathcal{O}\left ( \frac{(\log n)^{3/4}}{n^{1/4 + 2/(d+6)}} \right)\|f^q_{\ell-1}\|_\infty
\end{align*}
and when $d = 1$,
\begin{align*}
    \|&\mathbf{x}_{n,\ell}^p-\mathbf{P}_nf_\ell^p\|\\
    &\leq F_{\ell-1}\max_{1\leq q \leq F_{L-1}}\|\mathbf{x}^q_{n,\ell-1}-\mathbf{P}_nf^q_{\ell-1}\| \\
    &\quad+ F_{\ell -1}\tilde{C}\left (\mathcal{O}\left( \frac{\sqrt{\log n}}{n^{2/7}}\right )\|f^q_{\ell-1}\|_2+\mathcal{O}\left ( \frac{\sqrt{\log n}}{n^{1/2}}\right )\|f^q_{\ell-1}\|_\infty\right).
\end{align*}

Let $\epsilon_{n,\ell-1}=\max_{1\leq q \leq F_{L-1}}\|\mathbf{x}^q_{n,\ell-1}-\mathbf{P}_nf^q_{\ell-1}\|$ and define 
\begin{equation*}\delta_n=\begin{cases}
% \tilde{C}\left(\mathcal{O}\left(\frac{\sqrt{\log n}}{n^{2/(d+6)}}\right)\left(\underset{\ell, q}{\max} \|f^q_\ell\|+\mathcal{O}\left ( \left (\frac{\log n}{n} \right )^{1/4}\right)\underset{\ell, q}{\max}  \|f^q_\ell\|_\infty\right )\right)&\text{ if } d\geq 2\\
\tilde{C}\mathcal{O}\left(\frac{\sqrt{\log n}}{n^{2/(d+6)}}\right)\underset{\ell, q}{\max} \|f^q_\ell\| &\\
\qquad+\tilde{C}\mathcal{O}\left ( \frac{(\log n)^{3/4}}{n^{1/4 + 2/(d+6)}} \right)\underset{\ell, q}{\max}  \|f^q_\ell\|_\infty&\text{ if } d\geq 2\\
 % \tilde{C}\left(\mathcal{O}\left( \frac{\sqrt{\log n}}{n^{2/7}}\right )\underset{\ell, q}{\max} \|f^q_\ell\|+\mathcal{O}\left ( \left (\frac{\log n}{n}\right)^{1/2}\underset{\ell, q}{\max}  \|f^q_\ell\|_\infty\right)\right)&\text{ if } d=1
  \tilde{C}\mathcal{O}\left( \frac{\sqrt{\log n}}{n^{2/7}}\right )\underset{\ell, q}{\max} \|f^q_\ell\|&\\
  \qquad+\tilde{C}\mathcal{O}\left ( \frac{\sqrt{\log n}}{n^{1/2}}\right)\underset{\ell, q}{\max}  \|f^q_\ell\|_\infty&\text{ if } d=1
 \end{cases}.\end{equation*}
% and
% \[\tilde{\delta}_n = \kappa \left (\max\{C,1\}\mathcal{O}\left( \frac{\sqrt{\log n}}{n^{2/7}}\right )\max_{\ell,q} \|f^q_\ell\|+\mathcal{O}\left ( \left (\frac{\log n}{n}\right)^{1/2}\right )\max_{\ell,q} \|f^q_\ell\|_\infty\right).\]
Then, we have the recurrence relation
$$\epsilon_{n,\ell}\leq F_{\ell-1}(\epsilon_{n,\ell-1}+\delta_n.)%, \quad \epsilon_{n,0}=0,\qquad \tilde{\epsilon}_{n,\ell}=F_{\ell-1}\tilde{\epsilon}_{n,\ell-1}+\tilde{\delta}_n, \quad \tilde{\epsilon}_{n,0}=0
$$
Therefore, one may verify by induction that for all $\ell\geq 1$ we have 
$$ \epsilon_{n,\ell} \leq \delta_n\sum_{k=1}^{\ell} \prod_{j=\ell-k}^{\ell-1} F_j.%, \qquad  \tilde{\epsilon}_{n,\ell} \leq \tilde{\delta}_n\sum_{k=0}^{\ell-1} \prod_{j=\ell-k}^{\ell-1} F_j.
$$

The proof now follows from the definitions of $\epsilon_{n,\ell}$ and $\delta_n.$
    
\end{proof}

\end{document}